\documentclass[sigconf,nonacm]{acmart}
\renewcommand\footnotetextcopyrightpermission[1]{}
\AtBeginDocument{%
  }

\setcopyright{acmlicensed}
\copyrightyear{2018}
\acmYear{2018}
\acmDOI{XXXXXXX.XXXXXXX}
\acmConference[Conference acronym 'XX]{Make sure to enter the correct
  conference title from your rights confirmation email}{June 03--05,
  2018}{Woodstock, NY}
\acmISBN{978-1-4503-XXXX-X/2018/06}

\usepackage{hyperref}
\begin{document}

\title{VAR-3D: View-aware Auto-Regressive Model for Text-to-3D Generation via a 3D Tokenizer}

\author{Zongcheng Han}
%\email{20245227133@stu.suda.edu.cn}
\affiliation{%
  \institution{School of Computer Science and Technology}
  \state{Soochow University}
  \city{Suzhou}
  \country{China}
}

\author{Dongyan Cao}
\affiliation{%
  \institution{School of Computer Science and Technology}
  \state{Harbin Institute of Technology}
  \city{Harbin}
  \country{China}
}

\author{Haoran Sun}
\affiliation{%
  \institution{School of Computer Science and Technology}
  \state{Soochow University}
  \city{Suzhou}
  \country{China}
}

\author{Yu Hong}
\authornote{Corresponding author.}
\affiliation{%
  \institution{School of Computer Science and Technology}
  \state{Soochow University}
  \city{Suzhou}
  \country{China}
}


\begin{abstract}
Recent advances in auto-regressive transformers have achieved remarkable success in generative modeling. However, text-to-3D generation remains challenging, primarily due to bottlenecks in learning discrete 3D representations. Specifically, existing approaches often suffer from information loss during encoding, causing representational distortion before the quantization process. This effect is further amplified by vector quantization, ultimately degrading the geometric coherence of text-conditioned 3D shapes. Moreover, the conventional two-stage training paradigm induces an objective mismatch between reconstruction and text-conditioned auto-regressive generation. To address these issues, we propose View-aware Auto-Regressive 3D (VAR-3D), which intergrates a view-aware 3D Vector Quantized-Variational AutoEncoder (VQ-VAE) to convert the complex geometric structure of 3D models into discrete tokens. Additionally, we introduce a rendering-supervised training strategy that couples discrete token prediction with visual reconstruction, encouraging the generative process to better preserve visual fidelity and structural consistency relative to the input text. Experiments demonstrate that VAR-3D significantly outperforms existing methods in both generation quality and text-3D alignment.
\end{abstract}




\maketitle
\section{Introduction}
High-quality 3D generation is a fundamental challenge in computer vision, graphics, gaming, and immersive augmented and virtual reality systems. While auto-regressive transformers have achieved remarkable success in Large Language Models (LLMs) \cite{achiam2023gpt,tom20fewshot,chowdhery2023palm,ouyang2022training,touvron2023llama} and 2D generation \cite{sun2024autoregressive,tang2023make,yulanguage}, and Multimodal Large Language Models (MLLMs) \cite{alayrac2022flamingo,driess2023palme,sun2023emu}, extending these advances to text-to-3D synthesis remains non-trivial. This difficulty stems from the complex structural nature of 3D data and the stringent requirement for global spatial consistency when translating sparse textual descriptions into dense geometric representations. 

Recent research has explored diverse paradigms for 3D generation. Reconstruction-based approaches \cite{lrm24,tochilkin2024triposrfast3dobject} often struggle to infer plausible geometry from a single image, while methods leveraging 2D diffusion priors \cite{hgm,imagedream23,liu2023syncdreamer,mvdream23} are frequently limited by view inconsistency across synthesized images. Then, native 3D generative models \cite{jun2023shape-e,lan2024ln3diff,tang2024lgm,chen2024sar3d,zhang2025tar3d,siddiqui2023meshgpt,chen2024meshanything} have shown strong potential, supporting both image and text. Nevertheless, these approaches still face challenges in representation efficiency, scalability, and semantic fidelity between generated shapes and input text. Among existing methods, the combination of Vector Quantized-Variational AutoEncoder (VQ-VAE) \cite{Oord2017NeuralDR} and auto-regressive transformers has emerged as a practical framework for text-to-3D generation \cite{chen2024sar3d,zhang2025tar3d}. By mapping continuous geometric latent spaces into a discrete token, it reformulates 3D generation as a sequence modeling task, thereby leveraging the scaling potential of transformers.

However, an often overlooked issue is that the quantization process often acts as a filter, where structural topologies are collapsed into indices, making it difficult for the subsequent auto-regressive model to reconstruct fine-grained geometric details from a fragmented latent space. These approaches often suffer from excessive information compression and structural misalignment during encoding and quantization, degrading geometric coherence and limiting the ability of discrete tokens to capture fine-grained, text-specified details. Furthermore, the conventional two-stage paradigm creates a modality gap between representation learning and generative modeling. Since the auto-regressive transformer is optimized solely to minimize cross-entropy in the discrete token space, it remains ‘blind’ to the visual outcome. Consequently, even a high-likelihood token sequence can result in a  incoherent shape.

To address these issues, we propose View-aware Auto-Regressive 3D (VAR-3D), a unified framework for text-to-3D generation. First, we introduce a view-aware 3D VQ-VAE that captures complex geometric information, enabling its quantized tokens to better represent object features and improving data reconstruction quality. %VAR-3D improves discrete representations via view-aware interaction and multi-scale feature fusion, enhancing structural consistency across scales. 
Second, We further adopt a visual supervision training paradigm that aligns discrete sequence modeling with rendering-supervised, promoting geometric coherence and text–shape alignment. Extensive experiments validate the effectiveness of VAR-3D across multiple evaluation metrics. Our key technical contributions are summarized as follows:

\begin{itemize} 
\item We design a view-aware 3D VQ-VAE that effectively encodes intricate geometric structures, allowing its discrete tokens to more accurately represent object features and enhance reconstruction fidelity. 
\item We introduce a rendering-supervised training strategy that bridges the gap between discrete token prediction and visual reconstruction, effectively mitigating the representation-generation mismatch in text-to-3D tasks. 
\item Experiments demonstrate that VAR-3D outperforms existing approaches in text-guided 3D synthesis, delivering superior visual fidelity and geometric coherence.
\end{itemize}

\section{Related Works}
\subsection{3D Representation and Tokenization}
The effectiveness of 3D generative modeling largely depends on the underlying 3D data representation, which directly affects the fidelity, efficiency, and scalability of the generation process. Variational AutoEncoder (VAE) \cite{kingma2013vae} and their discrete variant, VQ-VAE \cite{Oord2017NeuralDR}, provide an effective framework for compressing high dimensional 3D data into compact latent spaces. Recent large-scale 3D generation methods \cite{lan2024ln3diff,wu2024direct3d,zhang2024clay} have further adapted continuous VAE representations to enable diffusion-based priors.

However, continuous latents often exhibit a trade-off between reconstruction fidelity and generative stability when used for powerful generative models. Discrete tokenization alleviates this issue by constraining the latent space and simplifying sequence modeling. VQ-VAE achieves this through a learnable codebook that discretizes continuous features, and has shown strong performance in generative tasks such as text-to-image synthesis \cite{ramesh2021zero} and other cross-modal applications \cite{ao2022rhythmic}. VQ-GAN \cite{esser2021taming} further improves perceptual quality through adversarial training.
Extending vector quantization to 3D remains non-trivial due to the view-dependent and multi-view consistent nature of 3D objects. Existing methods often employ view-agnostic quantization, which may discard important geometric and structural information. These limitations suggest that effective 3D discrete representations should explicitly account for cross-view interactions and hierarchical geometric information.
%This motivates the development of 3D-aware discrete representations that better preserve spatial structure.
\subsection{Paradigms in Text-to-3D Generation}
The generation of text into 3D has gone through several different stages of development, transitioning from optimization-based methods to feed-forward generative frameworks. Early Score Distillation Sampling (SDS) approaches \cite{chen2023fantasia3d,tang2023make,poole2022dreamfusion,chen2024comboverse,wang2024prolificdreamer} synthesize 3D content by distilling priors from pretrained 2D diffusion models. Although these methods are effective, they are limited by high computational cost, the “Janus” (multi-face) problem, and visual over-saturation, which restrict their scalability and stability in practice.

To improve efficiency, subsequent works \cite{long2023wonder3d,shi2023zero123plus,mvdream23,wang2024phidias,liu2025unidream} adopt two-stage pipelines that first generate consistent multi-view images, followed by fast 3D reconstruction \cite{lrm24,tang2024lgm,xu2024instantmesh}. Although significantly faster, these approaches remain constrained by the view-consistency of 2D generators, often leading to geometric artifacts such as floaters and fragmented structures. More recently, native 3D generative models \cite{lan2024ln3diff,lan2024gaussianything,zeng2022lion,li2024craftsman,3dshape2vecset,wang2023rodin,zhang2024clay} have emerged to directly model 3D distributions. Most existing methods in this category are based on diffusion formulations, which achieve high-quality generation but rely on slow iterative denoising processes. In parallel, auto-regressive models \cite{chen2024sar3d,zhang2025tar3d,wei2025octgpt} explore sequential or hierarchical token prediction for 3D generation. However, compared to diffusion-based methods, autoregressive models have been relatively less explored in the field of text-to-3D generation. Our VAR-3D further explores the potential of autoregressive models in the field of 3D generation.

\subsection{Generative Auto-Regressive Models}
Auto-regressive (AR) transformers have become a prominent paradigm for visual generation by modeling images or shapes as sequences of discrete tokens. Early works such as PixelCNN~\cite{Salimans2017PixeCNN} and VQ-GAN demonstrated that learning conditional distributions over visual tokens can produce high-quality results with strong global coherence. To alleviate the sequential bottleneck and improve scalability, subsequent studies have explored more efficient tokenization and generation strategies. For example, RQ-VAE~\cite{lee2022residualvq} employs residual quantization to capture fine-grained details, while the visual auto-regressive framework~\cite{tian2024var} reformulates next-token prediction into next-scale prediction, thereby achieving improved visual fidelity and accelerating sampling speed.

However, extending these successful approaches~\cite{chang2022maskgit, muse23, mar24, llamagen24} to text-to-3D generation remains challenging. Existing 3D autoregressive methods~\cite{siddiqui2023meshgpt, wang2024llamamesh, chen2024meshxl}, including those based on mesh tokenization, typically suffer from slow generation speed and insufficient robustness when handling geometric shapes. Furthermore, the widely adopted two-stage training paradigm, which separately optimizes the VQ-VAE and the AR transformer, introduces a mismatch between representation learning and sequence modeling. This discrepancy restricts the transformer’s ability to recover information lost during quantization, particularly for fine-grained text–shape alignment, highlighting the need for more coupled representation and training strategies in 3D auto-regressive modeling.

\begin{figure*}[h]
    \centering
    \includegraphics[width=\textwidth]{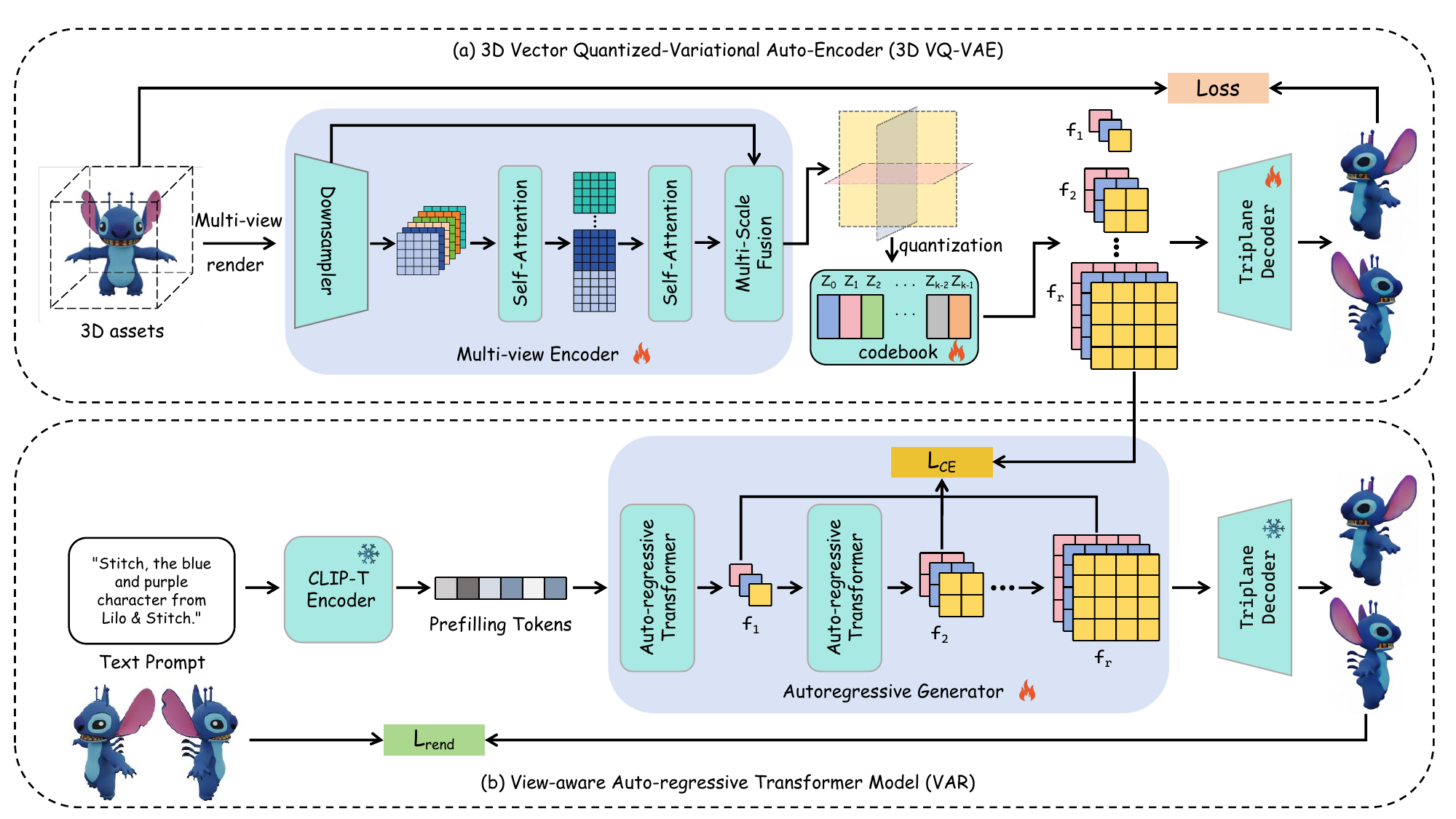}
    \caption{Overall architecture of the proposed VAR-3D framework. (a) 3D VQ-VAE: Multi-view renderings obtained from the 3D asset are encoded with dual self-attention to enhance feature interaction across views. Subsequently, the multi-view features are fused with multi-scale features from the downsampling stage and quantized to obtain latent triplane features $F=(f_1,\dots,f_r)$, which are decoded to reconstruct the 3D asset by a triplane decoder.
    (b) VAR model: In text-to-3D generation, the text is encoded into text features by CLIP-T. These text features are used as prefilling tokens for text conditional generation. Then the token sequence is progressively predicted across scales, and triplane features are synthesized via a codebook lookup to generate the 3D asset. Besides, a frozen triplane decoder provides visual supervision during training.}
    \label{fig:pipline}
\end{figure*}

\section{Methodology}
As illustrated in Figure~\ref{fig:pipline}, our framework consists of two primary components: a 3D VQ-VAE and a VAR model. In the first stage, the 3D VQ-VAE encodes 3D models into quantized triplane features, which are then represented as discrete multi-scale tokens. %These tokens serve as the training objective for the subsequent stage. 
In the second stage, the auto-regressive model performs next-scale prediction conditioned on text prompts. %%Regarding the optimization objective, we employ a dual-loss mechanism. 
In addition to the standard cross-entropy loss, we introduce rendering supervision. Specifically, the predicted tokens are reconstructed by 3D VQ-VAE deocder into 3D models to generate rendered images, which are used to provide additional geometric and visual constraints during training.

\subsection{3D VQ-VAE}
As shown in prior work \cite{lan2024ln3diff,chang2022maskgit,videoworldsimulators2024,rombach2022high}, high-quality visual generation relies on learning a compact and expressive latent space, typically achieved by a carefully designed variational autoencoder. Then, we introduce a 3D VQ-VAE that maps a 3D object into a discrete latent representation for sequence modeling under textual conditions.

\paragraph{\textbf{Encoder.}}As illustrated in Figure~\ref{fig:pipline}, given a 3D object, we first render a set of multi-view RGB-D images, denoted as S, where $S = \{\, s_{1}, s_{2}, \dots, s_{i} \,\}$. For each rendering $s_{i} = (I, D, C)$ , $I \in \mathbb{R}^{H \times W \times 3}$ represents the RGB image, $D \in \mathbb{R}^{H \times W}$ represents the depth map, and $C$ represents the corresponding camera parameters. To be compatible with both RGB images and depth maps, we transform the camera parameters $C$ into Plücker coordinates \cite{sitzmann2021lfns} to obtain $P \in \mathbb{R}^{H \times W \times 6}$. In this way, we obtain $s'_{i} = (I, D, P)\in \mathbb{R}^{H \times W \times (3+1+6)}$ by a concatenation operation.

To encode our input $s'$, we use a multi-view convolutional encoder \cite{mvdream23,tang2024lgm,chen2024sar3d}. For effective feature integration, we preserve view-aware interaction mechanism during the encode process. Specifically, we utilize the downsampling result $H =\{ h_i \}^6_{i=1} $ and apply self-attention mechanism on each view $h_i$ to capture details:
\begin{equation}
h_i = \text{SelfAttn}(h_i).\label{eq:1}
\end{equation}
Moreover, in order to allow one perspective to perceive other perspectives, we employ a cross-view attention mechanism to model dependencies between different perspectives. Specifically, we concatenate the multi-view features into a unified sequence and process them through a self-attention layer:
\begin{equation}
    H' = \text{SelfAttn}(\text{Concat}(H)),\label{eq:2}
\end{equation}
this mechanism ensures a comprehensive representation of the 3D structure by aggregating information across all available views.

To mitigate the loss of geometric awareness caused by feature compression during downsampling, we propose a multi-scale feature fusion strategy. We do not rely solely on the final bottleneck layer. Instead, we integrate hierarchical features $M$ from the last three downsampling layers into the latent representation $H''$:
\begin{equation}
    H'' = \text{Multi-scaleFusion}(H’,M),\label{eq:3}
\end{equation}
this multi-level integration effectively preserves both global structure and geometric details. Specifically, we uniformly interpolate the features of each layer of the hierarchical representation $M$ to match the scale of the latent representation $H'$, and then aggregate them through summation. %As a result, our approach establishes a more robust and high-fidelity 3D latent space. 
After this, we obtained the latent triplane \cite{lan2024ln3diff,wu2024direct3d,chen2024sar3d} features $f \in \mathbb{R}^{h \times w \times 3 \times d}$, where $h$ and $w$ represent the height and width of the triplane feature map, respectively, and $d$ represents the number of latent channels.

\paragraph{\textbf{Quantizer.}}We use the embedding vector $z_{q}$ from a learnable discrete codebook $Z = \{ z_{v} \}\ ^V_{v=1} \in \mathbb{R}^{d_q}$ to represent the continuous triplane features. Where $V$ is the size of the codebook. Specifically, we first need to interpolate the latent triplane features $f$ to obtain features maps $F =\{\ f_{r_1},f_{r_2},...,f_{r_k} \}\ $ at different scales $R=(r_1, r_2,...,r_k)$, which contributes to a progressive and coherent multi-scale representation. To ensure consistency across scales, we employ a shared codebook $Z$.  Then, we quantize the features $f_{r_i}\in \mathbb{R}^{(h_{r_i} \times w_{r_i} \times 3) \times d_q}$ at each scale to find the closest codebook entry $z_q$, as follows:
\begin{equation}
z_q := ( \arg\min_{z_v \in Z} \| {f_{r_i}} - z_v \| ) \in \mathbb{R}^{(h_{r_i} \times w_{r_i} \times 3 )\times d_q} \label{eq:4}.
\end{equation}

\paragraph{\textbf{Decoder.}}Following previous works \cite{wu2024direct3d,chen2024sar3d}, our decoder $D$ transforms discrete-scale triplanes into a triplane representation. The process begins by converting the discrete-scale tokens into latent features. Next, we perform scale unification to align these features into a consistent space. This results in a unified triplane feature, which is then processed by an upsampler to generate the final triplane representation. Finally, we render multiple views from this triplane to calculate the reconstruction loss.

\paragraph{\textbf{Training Objective.}}To achieve a stable optimization process while preserving high-fidelity geometry and appearance for mesh extraction, we adopt a hybrid training objective that jointly supervises volumetric rendering, discrete latent quantization, and perceptual realism. Specifically, the loss function is defined as:
\begin{equation}
L = \lambda_{\text{render}}L_{\text{render}} + \lambda_{\text{VQ}}L_{\text{VQ}} + \lambda_{\text{GAN}}L_{\text{GAN}},
\label{eq:5}
\end{equation}
where each term is weighted by a corresponding coefficient to balance its contribution during training.

\textit{Rendering Loss.} The rendering loss $L_{\text{render}}$ enforces consistency between the rendered outputs and the ground-truth RGB-D observations, including object silhouettes. To comprehensively capture both low-level reconstruction accuracy and perceptual similarity, we combine pixel-wise and perceptual metrics:
\begin{equation}
L_{\text{render}} = \lambda_{\text{MAE}}L_{\text{MAE}} + \lambda_{\text{SSIM}}L_{\text{SSIM}} + \lambda_{\text{LPIPS}}L_{\text{LPIPS}},
\label{eq:6}
\end{equation}
where Mean Absolute Error (MAE) penalizes pixel-wise discrepancies, Structural Similarity Index Measure (SSIM) preserves structural coherence, and Learned Perceptual Image Patch Similarity (LPIPS) \cite{zeng2022lion} encourages perceptual alignment in a learned feature space. %This multi-term formulation enables robust supervision for both geometry and appearance reconstruction under volumetric rendering.

\textit{Vector Quantization Loss.} To learn compact and discrete latent representations, we employ the vector quantization loss $L_{\text{VQ}}$ \cite{Oord2017NeuralDR} to supervise the update of the codebook. This loss function aims to minimize the difference between the encoder features and their corresponding quantized embeddings:
\begin{equation}
L_{\text{VQ}} = \sum_{r_i} \left( | \text{sg}[f_{r_i}] - z^{r_i}_q | + \beta | f_{r_i} - \text{sg}[z^{r_i}_q] | \right),
\label{eq:7}
\end{equation}
where $f_{r_i}$ denotes the encoder feature at region $r_i$, $\beta$ is the ratio between the two losses, $z^{r_i}_q$ represents the selected codebook vector, and $\text{sg}[\cdot]$ is the stop-gradient operator \cite{bengio2013estimating}. The first term is used to update the codebook, while the second term, the commitment loss, regularizes the encoder output to close the quantized embeddings, thus stabilizing the learning process of the discrete latent space.

%\paragraph{Adversarial Loss.} 
%We incorporate an adversarial loss $L_{\text{GAN}}$ \cite{gulrajani2017gans} to improve perceptual realism. The discriminator $D$ is optimized with WGAN-GP:
%\begin{equation}
%L_D = \mathbb{E}^g_{\tilde{x}}[D(\tilde{x})] - \mathbb{E}^r_{x}[D(x)] 
%+ \lambda \, \mathbb{E}_{\hat{x}}\big[(\lVert \nabla_{\hat{x}} D(\hat{x})\rVert_2 - 1)^2\big],
%\end{equation}
%where $x$ and $\tilde{x}$ denote ground truth and generated samples, respectively, and $\hat{x}$ is sampled by linear interpolation between them.  The generator minimizes
%\begin{equation}
%L_{\text{GAN}} = - \mathbb{E}_{z}[D(G(z))],
%\end{equation}
%with $G(z)$ producing generated samples from input latent codes $z$.  The gradient penalty weight is controlled by $\lambda$, ensuring stable training and sharper visual details.
\textit{Adversarial Loss.} We incorporate an adversarial loss $L_{\text{GAN}}$ \cite{gulrajani2017gans} to improve perceptual realism.  
The critic $D$ is trained with WGAN-GP; the loss for the critic (discriminator) is:
\begin{equation}
\scalebox{0.9}{$
L_D = \mathbb{E}_{\tilde{x}\sim p_g}[D(\tilde{x})]
- \mathbb{E}_{x\sim p_r}[D(x)]
+ \lambda\,\mathbb{E}_{\hat{x}\sim p_{\hat{x}}}
\big[(\lVert \nabla_{\hat{x}} D(\hat{x})\rVert_2 - 1)^2\big]
$}, 
\end{equation}
where $x\sim p_r$ and $\tilde{x}\sim p_g$ denote ground-truth and generated samples respectively. %, and $\hat{x}$ is sampled by linear interpolation $\hat{x}=\epsilon x + (1-\epsilon)\tilde{x}$ with $\epsilon\sim\mathcal{U}(0,1)$.  
The generator minimizes the objective:
\begin{equation}
L_{\text{GAN}} = - \mathbb{E}_{z\sim p_z}[D(G(z))],
\end{equation}
with $G(z)$ producing generated samples from latent codes $z\sim p_z$. %The gradient penalty weight $\lambda$ controls the penalty strength \cite{gulrajani2017gans}, which helps stabilize training and encourages sharper visual details.

\subsection{View-aware Auto-regressive Model}
The discrete token maps obtained from the first stage are used to train a text-conditioned autoregressive shape generation model. Our model follows a next-scale prediction strategy \cite{tian2024var,chen2024sar3d,zhang2025ar}, generating a sequence of token maps $\{f_{r_1}, f_{r_2}, \dots, f_{r_k}\}$ progressively guided by input the prefilling tokens ,which is generated from text.

\paragraph{\textbf{Sequence Construction.}}Initially, 3D shapes are encoded into discrete triplane features using a 3D VQ-VAE. Each feature is represented as an index from a learned codebook. These indices are then serialized into a multi-scale sequence. Within the sequence, indices of each plane are arranged in raster-scan order. To preserve spatial correlation across planes, indices corresponding to the same spatial location in the three planes are placed consecutively. %This organization ensures that the model can effectively learn local and cross-plane dependencies.

\paragraph{\textbf{Text-Conditional Generation.}}Our approach differs from feed-forward 3D reconstruction methods \cite{lrm24,tang2024lgm}, which directly map an input image to a 3D representation. Instead, we focus on generating 3D shapes conditioned on text prompts. Text embeddings are extracted using a CLIP-T \cite{Radford2021LearningTV} ViT-L encoder and injected into the autoregressive model via cross-attention. This allows semantic information from text to guide the generation process. To provide global semantic context, we use the pooled CLIP text embedding as an initial conditioning token for the generation sequence. %%This design stabilizes generation and ensures alignment between the text prompt and the generated 3D structure, enabling flexible and multimodal synthesis.
\begin{figure*}[t]
    \centering
    %\colorbox{white}{%
    \includegraphics[width=0.99\linewidth]{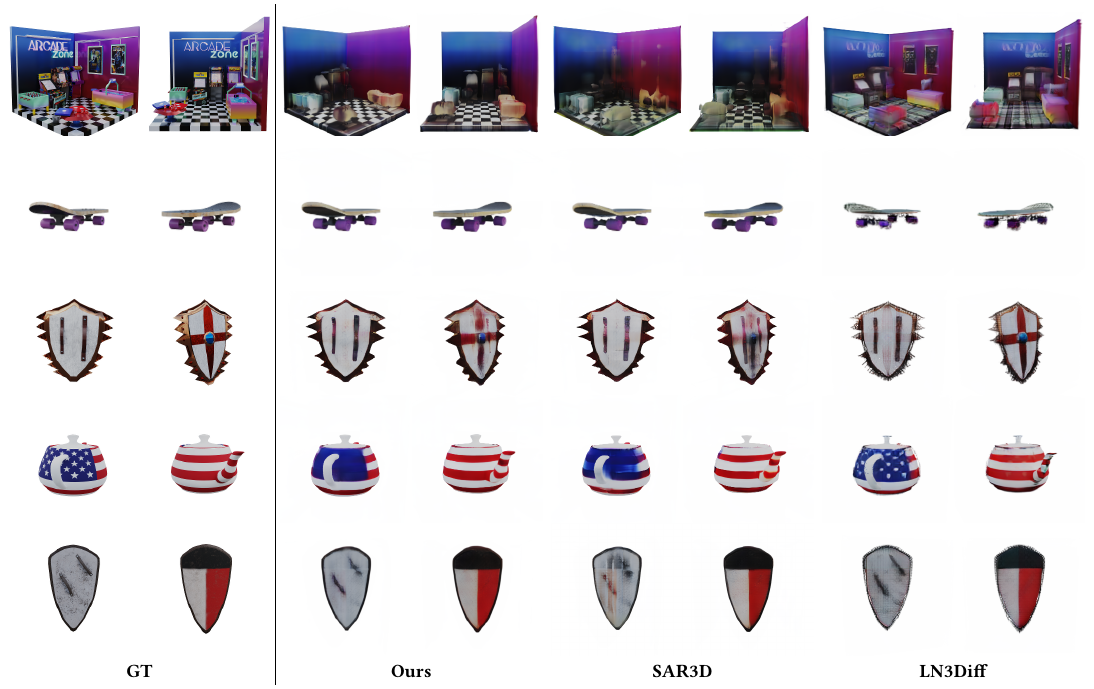}
    %}
    \caption{Visual comparison of 3D reconstruction. We present reconstruction 3D objects generated by our 3D VQ-VAE displaying two views of each sample. Compared to the baseline methods, our approach consistently yields better quality regarding geometry, texture. Furthermore, our method also demonstrates better performance in terms of multi-view consistency.}
    \label{fig:reconstruction_pdf}
\end{figure*}

\paragraph{\textbf{Autoregressive Generator.}}Generation proceeds from an initial low-resolution token map to progressively higher resolutions. The final token map $f_{r_k}$ matches the resolution of the original feature map. We implement the generator as GPT-2 style decoder-only Transformer \cite{tom20fewshot}. The autoregressive likelihood is formulated as:
\begin{equation}
p(f_{r_1}, f_{r_2}, \dots, f_{r_k}) = \prod_{i=1}^{k} p(f_{r_i} \mid f_{<r_i}),
\end{equation}
where $f_{<r_i}$ denotes all preceding scales. Tokens at each scale are predicted in parallel, while the model conditions on previous scales. For text-conditional synthesis, a text embedding $c$ acts as both the start token and a conditioning factor via Adaptive Layer Normalization (AdaLN) \cite{Peebles2022DiT}. Queries and keys are normalized to unit length before attention to improve training stability. %This approach allows the model to effectively incorporate semantic guidance at every resolution, facilitating fine-grained text–shape alignment. Meanwhile, the coarse-to-fine autoregressive formulation enables the model to capture global structure at lower resolutions and progressively refine local details at higher scales. 
The model is trained with a token-level cross-entropy loss:
\begin{equation}
L_{\text{CE}} = -\frac{1}{K} \sum_{i=1}^{K} \log p(f_{r_i} \mid f_{<r_i}, c),
\end{equation}
which encourages the model to assign high probabilities to the correct token maps and ensures accurate generation across all scales.

%\textbf{Rendering-Based Structural Supervision.}\; While $L_{\text{CE}}$ ensures token prediction accuracy, it does not explicitly enforce global geometric continuity or fine-grained structural details. To address this, we decode the predicted token maps using a pre-trained and frozen 3D VQ-VAE triplane decoder $D$ to reconstruct a 3D triplane representation. The reconstructed triplanes are projected into 2D images $I_{\text{pred}}$ from multiple views using a differentiable renderer $\mathcal{R}$. Ground-truth images $I_{\text{gt}}$ are rendered from the original 3D shapes under the same camera settings, and a rendering-based supervision is applied to encourage geometric consistency and  structural fidelity. %This supervision guides the model to maintain both structural consistency and visual fidelity.

%We define a multi-view rendering loss as a weighted combination of MAE and Mean Squared Error (MSE):
%\begin{equation}
%L_{\text{rend}} = \lambda_1 \| I_{\text{pred}} - I_{\text{gt}} \|_1 
%+ \lambda_2 \| I_{\text{pred}} - I_{\text{gt}} \|_2^2,\label{eq:9}
%\end{equation}
%where $\lambda_1$ and $\lambda_2$ balance the contributions of the two terms. This loss encourages the generated 3D shapes to better preserve visual appearance and structural consistency with the ground truth.
\paragraph{\textbf{Rendering-Based Structural Supervision.}}While $L_{\text{CE}}$ ensures token-wise prediction accuracy, it does not explicitly enforce global geometric continuity or fine-grained structural details. To bridge this gap, we introduce a differentiable rendering-based supervision pipeline. However, the standard discrete sampling (e.g., \textit{argmax}) of AR tokens is non-differentiable, which prevents gradients from the rendering loss from backpropagating to the AR model. To resolve this, we employ the Gumbel-Softmax Straight-Through Estimator (STE) to obtain a differentiable approximation of the discrete token map. Specifically, given the predicted logits $L$, we compute the differentiable one-hot representation $f_z$ as:
\begin{equation}
f_z = \text{Gumbel-Softmax}(L, \tau, \text{hard=True}),
\end{equation}
where $\tau$ is the temperature. The forward pass uses discrete selections to maintain fidelity, while the backward pass uses continuous relaxations to propagate gradients. These tokens are then mapped to the codebook and reshaped into a multi-scale 3D triplane representation. The reconstructed triplanes are then projected into 2D images $I_{\text{pred}}$ from multiple views using a differentiable renderer $\mathcal{R}$. Ground-truth images $I_{\text{gt}}$ are rendered from the original 3D shapes under the same camera settings. We define the multi-view rendering loss as a weighted combination of $\ell_1$ and $\ell_2$ losses:
\begin{equation}
L_{\text{rend}} = \lambda_1 | I_{\text{pred}} - I_{\text{gt}} |_1 + \lambda_2 | I_{\text{pred}} - I_{\text{gt}} |_2^2 ,\label{eq:9}
\end{equation}
where $\lambda_1$ and $\lambda_2$ are used to balance the contributions of each loss. %This joint optimization allows the rendering loss to provide direct semantic and structural guidance to the AR transformer, ensuring both geometric consistency and visual fidelity.

\paragraph{\textbf{Training Objective.}}The total loss used to train the autoregressive model combines token-level supervision and rendering-based visual supervision:
\begin{equation}
L_{\text{total}} = L_{\text{CE}} + \gamma L_{\text{rend}},\label{eq:10}
\end{equation}
where $\gamma$ balances the two training  objectives. 
The cross-entropy loss enforces accurate token prediction, while the rendering-based loss provides additional visual and geometric guidance. %, encouraging the predicted tokens to be consistent with coherent 3D structures across multiple resolutions.

%%The combination of multi-scale autoregressive modeling and rendering-based supervision allows the system to capture complex 3D structures while faithfully reflecting the input textual description.

\begin{table*}[ht]
\centering
\caption{Quantitative evaluation of text conditioned 3D generation. As shown below, the proposed method demonstrates strong performance across all metrics. KID scores are scaled by $10^2$, $\uparrow$: the higher value, the better performance, $\downarrow$: the lower the better.}
\label{tab:main experiment}
\begin{tabular}{clcc|ccccc|ccccc}
\toprule
&Method &&&& PSNR $\uparrow$ & SSIM $\uparrow$ & CLIP-T $\uparrow$ &&& FID $\downarrow$ & KID(\%) $\downarrow$ & LPIPS $\downarrow$ &
\\
\midrule
&Shape-E \cite{jun2023shape-e} &&&& 12.13 & 0.753 & 26.98 &&& 56.99 & 1.219 & 0.306 &\\
&LGM \cite{tang2024lgm}     &&&& 14.94 & 0.811 & 25.73 &&& 41.83 & 0.882 & 0.236& \\
&LN3Diff \cite{lan2024ln3diff} &&&& 15.70 & 0.767 & 25.84 &&& 61.16 & 1.523 & 0.292 &\\
&SAR3D \cite{chen2024sar3d}   &&&& 18.31 & 0.832 & \textbf{27.66} &&& 36.84 & 0.559 & 0.199&\\    
\midrule
&\textbf{Ours} &&&& \textbf{18.52} & \textbf{0.840} & 27.42 &&& \textbf{32.74} & \textbf{0.396} & \textbf{0.176}& \\
\bottomrule
\end{tabular}
\end{table*}

\section{Experiments}
\subsection{Experiment Setup}
\paragraph{\textbf{Datasets.}}We train our model on the G-Objaverse dataset \cite{qiu2023richdreamer,objaverse,objaverseXL}, %a large-scale rendered 3D asset dataset, where each 3D asset includes 
which contains RGB images, normal maps, depth maps, and corresponding camera poses. The dataset contains approximately 280K data %samples
spanning 10 general categories, including Human-Shape, Animals, Daily-Used, Furniture, Buildings \& Outdoor, Transportation, Plants, Food, Electronics, and Poor-quality. Following works \cite{lan2024ln3diff,chen2024sar3d}, we adopt a high-quality subset consisting of around 100K 3D assets for training. For text-conditioned generation, we use the captions as text prompts, which is provided by 3D-Topia \cite{chen20253dtopia,hong20243dtopia}.

\paragraph{\textbf{Metrics.}} We compare the novel views rendered from the synthesized 3D asset with the ground truth views based on a set of common metrics, including Peak Signal-to-Noise Ratio (PSNR), LPIPS \cite{zhang2018unreasonable}, SSIM \cite{wang2004image}. Additionally, we utlize Fréchet Inception Distance(FID) \cite{heusel2018fid}, and Kernel Inception Distance(KID) \cite{binkowski2018kid} to assess the overall distribution quality between the generated output and the ground truth, and use CLIP-T scores \cite{radford2021learning} to evaluate the consistency between the generated results and the input text prompts.

\paragraph{\textbf{Implementation Details.}} In our 3D VQ-VAE training stage, we use input images with a resolution of $H = W = 256$. The model takes 6 rendered views ($i = 6$) as multi-view inputs. We follow \cite{chen2024sar3d} by applying the feature map $R$ is quantized in a multi-scale manner across ten scales. We set the codebook size $V$ to 16,384 and the codebook channel dimension $d_q$ to 8. The hyperparameters in Eqn.\eqref{eq:5} are set as $\lambda_{\text{render}}=1$, $\lambda_{VQ}=1$ and $\lambda_{\text{GAN}}=0.025$, and Eqn.\eqref{eq:6} are set as $\lambda_{\text{MAE}}=1$ , $\lambda_{\text{SSIM}}=0.2$ and $\lambda_{\text{LPIPS}}=0.8$. We use a constant learning rate for training, initialized to $1e-4$ and adopt the AdamW optimizer [42] to train the 3D VQ-VAE.

To train our VAR model at second stage, our architecture is based on visual auto-regressive framework \cite{tian2024var}, adding plane positional encoding for each plane, which has 16 transformer blocks with 16 heads. We use the AdamW optimizer set the learning rate of $1e-4$, with a linear decaying learning rate schedule for training. At the same time, for rendering-based structural supervision, we randomly select a view rendering from 6 views as supervision and we use frozen triplane decoder to reonstruct pred model. The hyperparameters in Eqn.\eqref{eq:9} and Eqn.\eqref{eq:10} are set as $\lambda_{1} = 0.2$ , $\lambda_{2} = 8$ and $\gamma=0.1$. We set the batch size to 32 for 3D VQ-VAE with 100K iterations and to 16 for  VAR model with 100 epochs. The entire training procedure was performed using 2 NVIDIA RTX 4090 GPUs. %, with each stage training lasting 10 days.

\subsection{Reconstruction Result}
We first compare the reconstruction fidelity of our method with other triplane-based approaches. LN3Diff \cite{lan2024ln3diff} uses a continuous latent triplane representation with VAE, whereas SAR3D \cite{chen2024sar3d} adopts a discrete representation based on VQ-VAE. %Compared to these methods, our approach achieves higher reconstruction accuracy, highlighting the effectiveness of our discrete triplane encoding in capturing fine-grained geometry and appearance details.

For appearance evaluation, we compute the PSNR and SSIM between rendered reconstructions and ground-truth images. We further assess distribution-level quality using FID and KID. As shown in Table~\ref{tab:reconstruction result}, our method consistently outperforms all baselines across all metrics. Specifically, it achieves a PSNR of 28.97 and an SSIM of 0.938, surpassing LN3Diff $(25.63 / 0.894)$ and SAR3D $(28.04 / 0.927)$. At the distribution level, our method attains lower FID and KID scores (30.50 and 0.140), compared to $35.92 / 0.161$ for SAR3D and $63.35 / 1.044$ for LN3Diff, indicating our model better alignment with the ground truth data distribution.

Qualitative comparisons in Figure~\ref{fig:reconstruction_pdf} further validate these results. Our method produces reconstructions with more faithful appearances and finer geometric details, closely matching the ground-truth 3D models. %In contrast, LN3Diff \cite{lan2024ln3diff} retains partial details for complex objects but suffers from limited appearance and geometric accuracy, while SAR3D \cite{chen2024sar3d} generates relatively accurate geometry yet exhibits view inconsistencies that degrade overall reconstruction quality.
In contrast, while LN3Diff \cite{lan2024ln3diff} preserves some details of complex objects, it suffers from shortcomings in the appearance of object and geometric accuracy; while SAR3D \cite{chen2024sar3d} generates relatively accurate geometric shapes, it suffers from viewpoint inconsistency issues, thus reducing the overall reconstruction quality.

\begin{table}[t]
\centering
\caption{Quantitative comparison of 3D reconstruction. We evaluate the reconstruction fidelity based on different methods using latent triplane representation.}
\label{tab:reconstruction result}
\begin{tabular}{lc|cccc|ccc}
\toprule
Method & & &PSNR $\uparrow$ & SSIM $\uparrow$ &&& FID $\downarrow$ & KID(\%) $\downarrow $  \\ \midrule
LN3Diff \cite{lan2024ln3diff} &&& 25.63 & 0.894 &&& 63.35 & 1.044 \\
SAR3D \cite{chen2024sar3d}    &&& 28.04 & 0.927 &&& 35.92 & 0.161 \\ 
\midrule
\textbf{Ours} &&& \textbf{28.97} & \textbf{0.938} &&& \textbf{30.50} & \textbf{0.140}  \\
\bottomrule
\end{tabular}
\end{table}

\begin{figure*}[ht]
    \centering
    %\colorbox{white}{%
    \includegraphics[width=\textwidth]{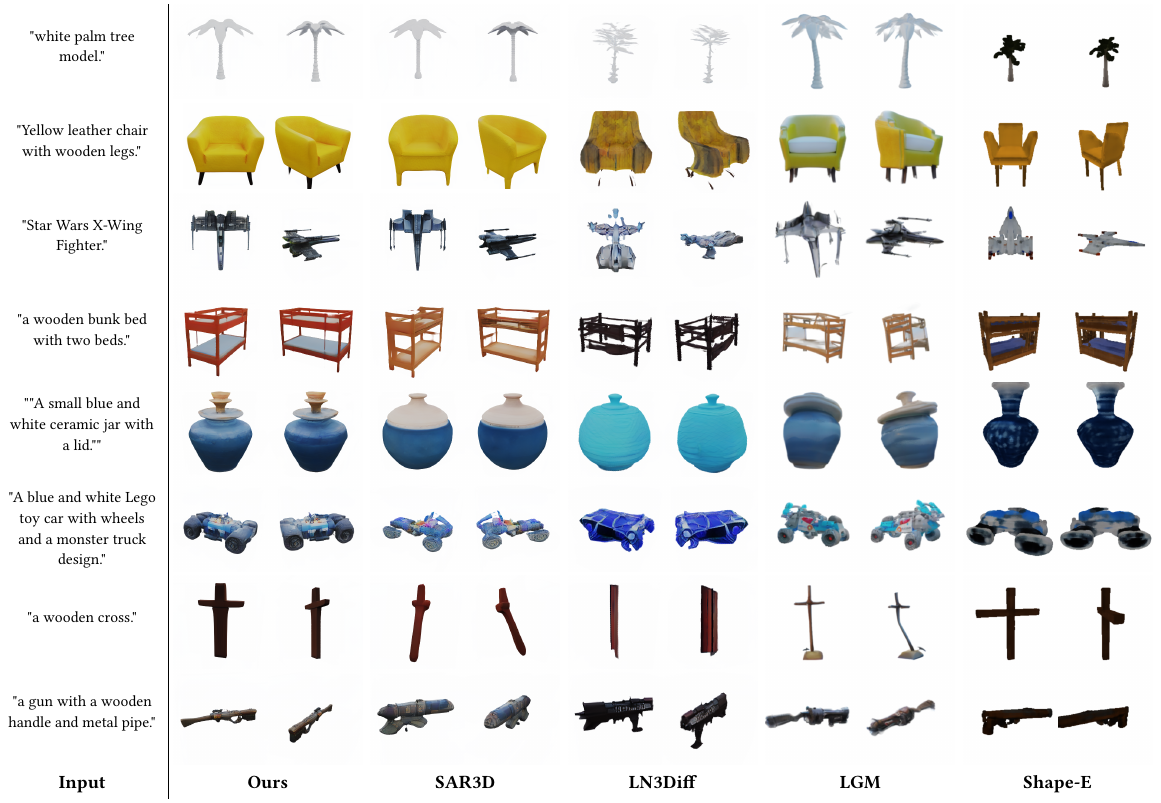}
    %}
    \caption{
    %Qualitative Comparison of text-conditioned 3D generation. We present text-to-3D generation results objects generated by our method, displaying two views of each sample. Compared to the baseline methods, our approach consistently yields better quality regarding geometry, texture and text-3D alignment.
    Qualitative comparison of text-conditioned 3D generation. We present text-to-3D generation results for objects generated by our method, displaying two views of each sample. Compared to the baseline methods, our approach consistently yields better quality regarding geometry, texture, and text–3D alignment.
    }
    \label{fig:main_comparison_pdf}
\end{figure*}

\subsection{Generation Result}
%In this section, we first conduct quantitative evaluations using standard metrics to compare our method with existing baseline approaches. Then, we provide qualitative results to visually demonstrate the fidelity, consistency, and structural details of the generated 3D content.
In this section, we first report quantitative comparisons with existing methods using standard evaluation metrics. Complementary qualitative results are provided to illustrate the visual quality, structural coherence, and consistency of the generated 3D content.

\paragraph{\textbf{Quantitative Evaluation.}}We performed a quantitative comparison of 500 test samples
from the Objaverse dataset \cite{objaverse,objaverseXL}. %, following the data classification method used in G-Objaverse.
For each 3D asset, a corresponding text prompt is generated using 3D-Topia, and all methods are evaluated by rendering the generated shapes from 6 views at a resolution of $256 \times 256$.

%As reported in Table~\ref{tab:main experiment}, our method demonstrates consistently strong performance across both reconstruction and perceptual metrics. Compared to existing approaches, our model achieves the highest PSNR (18.52) and SSIM (0.840), indicating superior reconstruction fidelity and more accurate geometric recovery. While SAR3D attains the best CLIP-T score (27.66), our method achieves a comparable CLIP-T of 27.42, reflecting strong text–3D semantic alignment. In terms of perceptual quality and distribution-level similarity, our approach outperforms all prior methods, achieving the lowest FID (32.74), KID (0.396), and LPIPS (0.176) scores. These results suggest that the generated 3D assets are not only closer to the ground-truth distribution but also exhibit improved perceptual realism. Overall, our method achieves a favorable trade-off across reconstruction accuracy, perceptual quality, and text-conditioned alignment.
As reported in Table~\ref{tab:main experiment}, our method demonstrates strong and consistent performance across both reconstruction and perceptual metrics. It achieves the highest PSNR (18.52) and SSIM (0.840) among all methods, indicating improved reconstruction fidelity and more accurate geometric recovery. Although SAR3D attains the best CLIP-T score (27.66), our method achieves a comparable CLIP-T of 27.42, suggesting similarly strong text–3D semantic alignment. In addition, our approach outperforms prior methods in terms of perceptual quality and distribution-level similarity, achieving the lowest FID (32.74), KID (0.396), and LPIPS (0.176). These results indicate that our method produces 3D assets that are closer to the ground-truth distribution while maintaining high perceptual realism, leading to a favorable trade-off between reconstruction accuracy, perceptual quality, and text-conditioned alignment.

\paragraph{\textbf{Qualitative Evaluation.}}Figure~\ref{fig:main_comparison_pdf} presents a qualitative evaluation of our method against representative text-conditioned 3D generation approaches, including 2D-assisted methods LGM \cite{tang2024lgm}, native 3D diffusion models: Shape-E \cite{jun2023shape-e}, LN3Diff, and autoregressive 3D generation methods SAR3D. Meanwhile, it also proved that our quantitative assessment was correct and effective. 
\begin{table}[t]
\centering
\caption{Ablation study of different components in our 3D VQ-VAE model. We ablate the design of our model architecture. Each component contributes to consistent gains in reconstruction performance and improves the modeling capacity.}
\label{tab:vae ablation}
\begin{tabular}{l|ccc}
\toprule
Setting & PSNR $\uparrow$ & FID $\downarrow$ & KID(\%) $\downarrow $  \\ \midrule
base & 28.42 & 34.72 & 0.165 \\
w/ view-aware interaction   & 28.57 &  33.35 & 0.151 \\ 
w/ multi-scale fusion & 28.68  & 32.00 & 0.150 \\
\midrule
full & 28.97 & 30.50 & 0.140  \\
\bottomrule
\end{tabular}
\end{table}

%As shown in Figure~\ref{fig:main_comparison_pdf}, Our method outperforms previous approaches, offering not only more vivid appearances and finer geometries but also more precise alignment with the provided text prompts. Furthermore, other methods suffer from varying degrees of quality degradation. For native 3D diffusion methods, the generated objects easily lack distinctive features and geometric details, and are limited by the finite reconstruction accuracy of their latent representations. For example, the object generated by LN3Diff with the prompt:\textit{"A small blue and white ceramic jar with a lid."} exhibit inherent flaws. Although LGM can generate plausible objects, it fails to provide reasonable geometric shapes, which is an inherent limitation of their 3D Gaussian representation. Furthermore, while SAR-3D can generate images of good quality, there may be instances where the results do not match the given text description. Like the case of \textit{"Yellow leather chair with wooden legs"}, the model fails to understand the \textit{"wooden legs"} feature.
As shown in Figure~\ref{fig:main_comparison_pdf}, our method consistently outperforms previous approaches by producing results with more vivid appearances, finer geometric details, and more accurate alignment with the provided text prompts. In contrast, existing methods suffer from varying degrees of quality degradation in both geometry and semantic consistency. Specifically, native 3D diffusion methods tend to generate objects that lack distinctive features and fine-grained geometric details, largely due to the limited reconstruction accuracy of their latent representations. For instance, LN3Diff fails to faithfully capture structural details in the example prompted by \textit{"A small blue and white ceramic jar with a lid."}, resulting in visible geometric artifacts. Although LGM is able to generate visually plausible objects, it often struggles to produce reasonable and coherent geometric shapes, which can be attributed to the inherent limitations of its 3D Gaussian representation. Furthermore, while SAR3D can synthesize images with relatively good visual quality, its results may not always align well with the given text descriptions. As illustrated by the prompt \textit{"Yellow leather chair with wooden legs."} SAR3D fails to correctly capture the \textit{"wooden legs"} attribute, indicating insufficient text–geometry understanding.

\subsection{Ablation Studies}
We conduct ablation studies to analyze the contributions of key components in our framework. Specifically, we evaluate the effects of the multi-scale feature fusion and view-aware interaction modules on the 3D VQ-VAE, examine the influence of different codebook sizes during training, and assess the impact of vision-based supervision strategy on the autoregressive model.

\paragraph{\textbf{3D VQ-VAE Ablations.}}We assess the effectiveness of the proposed view-aware interaction and multi-scale fusion modules %through an ablation study on the VQ-VAE trained with 20k samples for 100K steps
, as summarized in Table~\ref{tab:vae ablation}. 
%Removing either component consistently degrades performance across all metrics. 
Specifically, disabling multi-scale fusion reduces the PSNR from 28.97 to 28.57 and increases the FID to 33.35 and the KID to 0.151, while removing the view-aware interaction module also results in lower PSNR and higher FID/KID scores. As shown in Table~\ref{tab:codebooksize}, changing the codebook size in the 3D VQ-VAE %model
significantly affects model performance. A larger codebook yields better reconstruction results, indicating that the model is better able to capture shape diversity and preserve a wider variety of shapes.

%Qualitative results in Figure~\ref{fig:ablation_pdf} further support these observations. Without these modules, the model suffers from geometric distortions and diminished fine details under multi-view interference. In contrast, their inclusion significantly enhances the preservation of structural integrity and local details, yielding reconstructions that are markedly closer to the ground truth. Overall, the full model achieves the best performance, demonstrating that view-aware interaction and multi-scale fusion contribute to a more robust and expressive latent representation.

\begin{figure}[t]
    \centering
    \includegraphics[width=\linewidth]{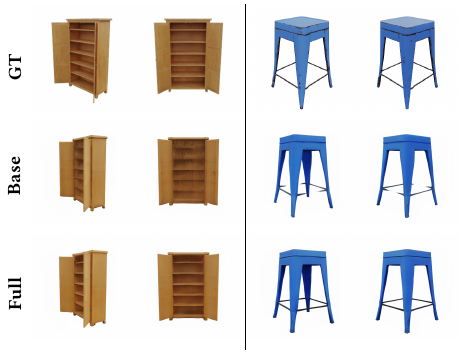}
    \caption{Visual comparison of our 3D VQ-VAE ablation experiments on the effectiveness of our components}
    \label{fig:ablation_pdf}
\end{figure}

Qualitative results in Figure~\ref{fig:ablation_pdf} provide further evidence supporting these findings. When the proposed modules are removed, the model exhibits noticeable geometric distortions and struggles to maintain fine-grained details, particularly under the interference introduced by multiple views. In contrast, incorporating the proposed modules substantially improves reconstruction quality, leading to better preservation of structural coherence and sharper local details across views, resulting in inconsistent geometry across views. %The resulting reconstructions exhibit clearer object boundaries, more consistent geometry, and closer visual alignment with the ground-truth shapes.

%\textbf{Training objective Ablations.}\; 
%We conduct ablation studies on the training objectives to investigate the effect of the proposed visual rendering-based loss ($L_{rend}$). Due to the autoregressive formulation of our model, the cross-entropy loss ($L_{CE}$) is required to provide token-level supervision and is therefore always retained. After training the model for 50 epochs, we compare the baseline setting using only the cross-entropy loss with the full objective that additionally incorporates the rendering-based loss.

%As shown in Table~\ref{tab:ar model ablation}, incorporating the rendering-based loss improves FID from 69.37 to 67.22 and reduces KID from $2.065$ to $1.968$, indicating a clearer alignment between the generated samples and real data distributions. In contrast, the CLIP-T score slightly decreases from 26.96 to 26.81, suggesting that the rendering-based perceptual supervision mainly enhances visual fidelity and distribution-level quality rather than semantic consistency. Overall, these results demonstrate that the proposed rendering-based loss provides effective complementary supervision to the cross-entropy objective in the autoregressive setting.
\begin{table}[t]
\centering
\caption{Ablation study of different codebook size in our 3D VQ-VAE model. Larger codebooks increase shape diversity and enhance reconstruction performance.}
\label{tab:codebooksize}
\begin{tabular}{cc|cccccc}
\toprule
Codebook Size  &&& PSNR $\uparrow$ & SSIM $\uparrow$ & LPIPS $\downarrow$ &\\ 
\midrule
4096   &&& 28.58 & 0.932 & 0.069 & \\
8192   &&& 28.87 & 0.936 & 0.065 & \\ 
16384  &&& 28.97 & 0.938 & 0.063 & \\
\bottomrule
\end{tabular}
\end{table}

\paragraph{\textbf{Training Objective Ablations.}}We conduct ablation studies on the training objectives to evaluate the impact of the proposed rendering-based visual loss ($L_{\text{rend}}$) in the second-stage autoregressive model. %Our full model is trained using the combined objective of cross-entropy loss ($L_{\text{CE}}$) and rendering-based loss ($L_{\text{rend}}$) for 100 epochs on the full training set.
To assess the contribution of $L_{\text{rend}}$, we adopt a removal-based ablation protocol.
Specifically, %starting from the pre-trained model, 
we fine-tune the autoregressive transformer under different loss configurations using a reduced subset of 20k training samples for 20 epochs. The considered settings include retaining the full objective ($L_{\text{CE}} + L_{\text{rend}}$) and removing the rendering-based loss, i.e., optimizing with $L_{\text{CE}}$ only.

This allows us to isolate the effect of the rendering-based supervision while keeping the model initialization, tokenizer, and token distribution fixed, and focuses on evaluating the performance degradation caused by removing $L_{\text{rend}}$. As shown in Table~\ref{tab:ar model ablation}, removing the rendering-based loss leads to a noticeable degradation in FID and KID, indicating that $L_{\text{rend}}$ plays an important role in maintaining visual fidelity and distribution-level quality.
Meanwhile, the CLIP-T score remains largely stable with only a %minor 
decrease, while FID and KID show clear improvements, indicating that the rendering-based loss enhances distribution-level visual quality without compromising text–3D semantic alignment. These results demonstrate that the rendering-based loss effectively complements cross-entropy, %and preserves generation quality.
enabling better learning of 3D discrete representations. 

\section{Limitations}
Although VAR-3D produces high-quality 3D objects, it still relies on a two-stage training pipeline rather than a fully end-to-end design. Future work may explore unified end-to-end 3D generation. Moreover, the current framework only supports text-conditioned generation and does not handle image or multimodal inputs. In addition, the quality of geometry and texture is constrained by volume rendering, and adopting more efficient 3D representations \cite{Huang2DGS2024} could further improve results. Finally, due to limited computational resources, we did not further explore the impact of visual supervision on the model or the scalability of our method. Nevertheless, prior work on 2D scaling laws in visual auto-regressive frameworks \cite{tian2024var} indicates its could potentially extend to larger-scale 3D content.

\begin{table}[t]
\centering
%\caption{Ablation study of training objective. Without visual supervision, the generated distribution shows a decline. }
\caption{Ablation study of training objectives at second stage.
Introducing visual supervision, $L_{\text{rend}}$ improves distribution-level similarity, as reflected by lower FID and KID.}
\label{tab:ar model ablation}
\begin{tabular}{ccc|cccc}
\toprule
&Training Objective &&& $L_{CE}$ & $L_{CE}+L_{\text{rend}}$\\ 
\midrule
&CLIP-T $\uparrow$     &&& 27.26 & 27.17 & \\ 
&FID $\downarrow$      &&& 42.61 & 41.26 & \\
&KID(\%) $\downarrow$  &&& 0.701 & 0.655 & \\
\bottomrule
\end{tabular}
\end{table}

\section{Conclusion}
In this work, we presented VAR-3D, a novel framework that advances 3D object generation by text-conditioned. By introducing the %multi-scale fusion and view-aware interaction in a 
view-aware 3D VQ-VAE, we mitigated compression losses and view inconsistencies during the encoding 3D objects,  %stage of,
thereby achieving high-quality 3D reconstruction and representation. Furthermore, we use the frozen 3D VQ-VAE decoder as a visual reconstruction supervision during the second stage of the training process, ensuring that the generated 3D objects conform not only to the latent triplane distribution but also to the vision-based distribution. Both quantitative and qualitative results demonstrate the effectiveness of our method, highlighting the potential of auto-regressive 3D generation. Future research could further explore end-to-end training and extend it to a wider range of 3D content, multimodal-condition generation and understanding challenges.

%\begin{acks}
%To Robert, for the bagels and explaining CMYK %and color spaces.
%\end{acks}
\bibliographystyle{ACM-Reference-Format}
\bibliography{main}
\appendix

\end{document}